\documentclass{article}
\pdfoutput=1
\usepackage{spconf}
\usepackage{amsmath}
\usepackage{graphicx}
\usepackage{algorithmic}
\usepackage{multirow}
\usepackage{algorithm}
\usepackage{color}
\usepackage{enotez}
\let\footnote=\endnote
\usepackage{tablefootnote}

\usepackage{enumitem}
\setlist{nosep, leftmargin=14pt}

\usepackage{mwe} 

\let\footnote=\endnote

\newcommand{\final}[1]{\textcolor{black}{#1}}

\title{Spinal Osteophyte Detection via Robust Patch Extraction on Minimally Annotated X-rays}
\name{ Soumya Snigdha Kundu$^{\star}$\thanks{*S.S.K carried out the work as a part of the Big Data Institute Summer Internship Programme at the University of Oxford.} \qquad Yuanhan Mo$^{\dagger}$ \qquad Nicharee Srikijkasemwat$^{\star\star}$ \qquad Bart\l omiej W. Papie\.z $^{\dagger}$}

\address{$^{\star}$ Department of Surgical \& Interventional Engineering, King's College London, London, UK \\
$^{\star\star}$ Department of Engineering Science, University of Oxford, Oxford, UK \\
    $^{\dagger}$ Big Data Institute, University of Oxford, Oxford, UK } 
\begin{document}
%
\maketitle


\begin{abstract}
The development and progression of arthritis is strongly associated with osteophytes, which are small and  elusive bone growths. This paper presents one of the first efforts towards automated spinal osteophyte detection in spinal X-rays.
A novel automated patch extraction process, called SegPatch, has been proposed based on deep learning-driven vertebrae segmentation and the enlargement of mask contours. 
A final patch classification accuracy of 84.5\% is secured, surpassing a baseline tiling-based patch generation technique by 9.5\%. 
This demonstrates that even with limited annotations, SegPatch can deliver superior performance for detection of tiny structures such as osteophytes.
The proposed approach has potential to assist clinicians in expediting the process of manually identifying osteophytes in spinal X-ray.
\end{abstract}

\section{Introduction}
\label{sec:intro}
One of the features observed in osteoarthritis (OA) \cite{van2007osteophytes} are osteophytes which are  fibrocartilage-capped  bony outgrowths. They play a strong role in determining the severity and progression of OA. Osteophytes also cause problems by impinging nerves such as spinal roots \cite{jones2012l5} and pose significant issues in many other spinal diseases.
\par
The manual identification of osteophytes has been generally done using
Magnetic Resonance Imaging (MRI) \cite{tozawa2021possible}. In contrast, the use of X-rays has been limited due to the challenges associated with the quality and 2D nature of conventional radiography. 
Automated detection of spinal osteophytes has also been explored. 
Wang et al. \cite{wang2016detection} extracted region proposals and classified them from unwrapped cortical shell maps in Positron Emission Tomography and Computed Tomography scans of the vertebral bodies of the thoracic and lumbar spine. 
They achieved 85\% sensitivity at 2 false positive detections per patient. 

Automated approaches to detect osteophytes in other parts of the body has been proposed as well. 
Ebsim et al., \cite{ebsim2022automatic} identified hip osteophytes in dual-energy X-ray absorptiometry by segmenting the osteophyte regions using U-Net \cite{ronneberger2015u} and manual cropping (to highlight the area of interest). 
Banerjee et al \cite{banerjee2011osteophyte} identified hand osteophytes in X-rays through cellular neural networks (neural networks where communication is allowed only between neighboring units). They automated the ground truth acquisition but they lacked any involvement  from an actual clinical representative throughout the acquisition process. 
\final{The VinDr-SpineXR study \cite{nguyen2021vindr} provides an extensive dataset with baseline results on multiple spinal diseases, including osteophytes. The poor results (and trend of increasing performance with increasing label size) match with our detection findings in sec 5.3.}


\begin{figure}
    \centering
    \includegraphics[width=0.48\textwidth]{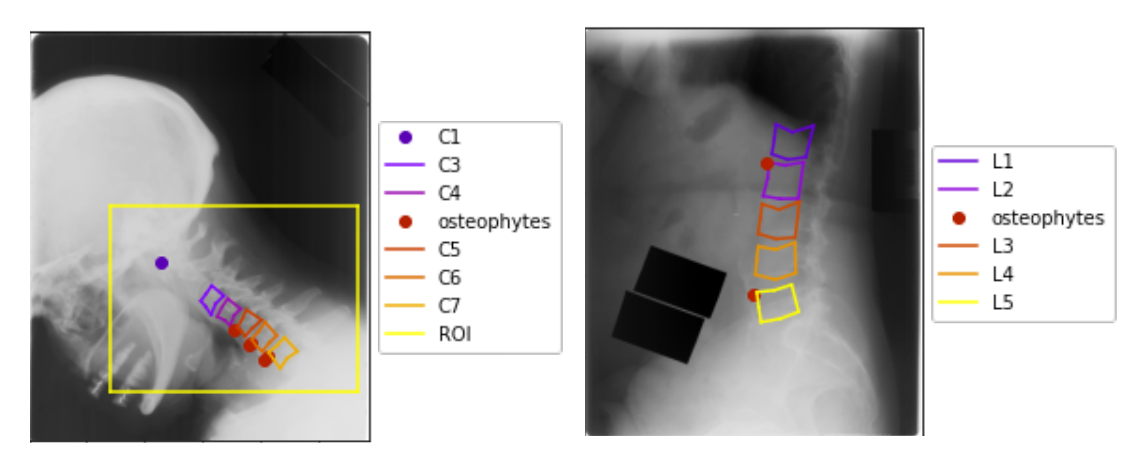}
    \caption{Example of annotations for cervical (left) and lumbar (right) X-ray scan available from NHANESII$^2$ data set.}
    \label{fig:nhanes-osteophyte-label}
\end{figure}

The number and size of the osteophytes could be seen as an estimate of the OA progress. Clinical experts' annotations are time-consuming and expensive while being prone to errors (i.e. inter- and intra-reader variability). Especially for tiny structures such as osteophytes. To the best of our knowledge, there has been no automated method for spinal osteophyte detection for spinal X-rays. 

\begin{figure*}[t]
    \centering
    \includegraphics[width=1\textwidth]{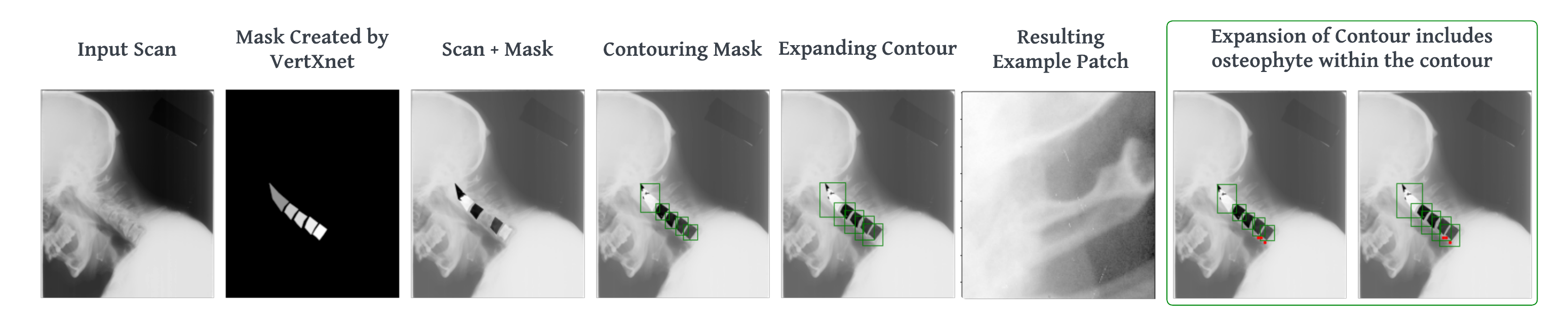}
    \caption{A representation of the SegPatch pipeline with a resulting patch. The green box highlights the necessity to expand contours to fully encompass the osteophytes within the boundary. Red dots denote the osteophytes.}
    \label{fig:pipeline}
\end{figure*}


Our contributions are as follows - We have developed a method to automatically identify osteophytes from conventional spinal radiography. First, we proposed a robust mechanism SegPatch, to extract patches from the spinal X-ray images through a deep-learning based vertebrae body segmentation approach \cite{chen2023vertxnet}.

Unlike the standard tiling approach, the newly proposed patching generates patches that are associated with potential locations of the osteophytes. Thereafter, we trained multiple neural network models for classification. 
Our framework was evaluated using the publicly available spine X-ray data set NHANESII. We compared SegPatch with the baseline tiling method. The results show the significant advantage of using SegPatch.


\section{Data}
The method was developed based on the publicly available data set\footnote{This research study was conducted retrospectively using human subject data made available in open access$^2$. Ethical approval was not required as confirmed by the license attached with the open access data.} collected via the Second National Health and Nutrition Examination Survey (NHANES II)\footnote{Dataset information can be found here: \newline
https://data.lhncbc.nlm.nih.gov/public/NHANES/X-rays/README.html .} conducted by the National Center for Health Statistics
during the years 1976-1980. 
The data encompasses digitized versions of the 17100 X-ray films. 9670 were cervical (of the size 1462 x 1755) and 7430 lumbar (of the size 2048 x 2487). 
However, only 241 cervical and 232 lumbar X-ray images were annotated for pixel coordinates indicating the presence of osteophytes. 
Each vertebrae was delineated using six pixel points.
A 75:25 split was employed for training and testing data.
The exemplar radiographs and the associated annotations have been shown in Fig.~\ref{fig:nhanes-osteophyte-label}.


\section{Methodologies}
Given the challenges associated with the detection of the miniscule osteophytes, the main goal of this research was to create a technique that could ascertain whether a particular image patch contained an osteophyte.
As seen in Fig.~\ref{fig:nhanes-osteophyte-label}, osteophytes generally tend to occur on the corners of the vertebrae. 
Based on this observation we developed a pipeline to extract several patches from high-resolution radiographs (Sec. \ref{sec:patch_creation}) and classify those patches for the presence of an osteophyte (Sec. \ref{sec:patch_classification}). 
Therefore, we set the hypothesis that if a patch is classified as "present", then either of the two corners of the vertebrae will have an osteophyte.
If the patch contains only one corner then it is bound to have a single osteophyte. 

For an qualitative evaluation, \final{the explainability of the classifier was tested via class activation map generation method SS-CAM \cite{wang2020ss}, to} understand and verify the focus of the classifier.


\subsection{Robust Patch Creation}
\label{sec:patch_creation}
Two methods of patch creation were investigated in this study.
\subsubsection{Tiling}
The tiling approach was inspired from Ozge et al.~\cite{ozge2019power}. The authors proposed it for a detection task in high-resolution pedestrian and vehicle images. Here, tiling is used as an image pre-processing step to split the image into equal-sized non-overlapping patches and label them based on whether a patch contained an annotation - indicating the presence of an osteophyte. 
To ensure the corner regions were not split amongst patches, the annotations were considered as boxes.
The tiling approach generated a highly imbalanced set of patches. The number of patches containing osteophytes amounted to 1453 ($<$ 5\%). Those without equalled to 32160. The boxes used as annotations were also used as the ground truth bounding boxes for the detection task in section \ref{detection}.

\subsubsection{SegPatch: Segmentation-driven Patch Extraction}
The main challenge in tiling was the significant class imbalance generated by considering the complete scan. Any approach which could focus on the vertebrae region alone, would potentially improve performance as the amount of noise seen by the classifier would reduce.
Therefore, a custom patch-generating method called SegPatch was devised to overcome the above limitations.

The SegPatch process begins by utilizing a vertebrae segmentation network VertXNet \cite{chen2023vertxnet} to localise the vertebrae in the scan. 
However, while the segmentation accounted for each vertebra in each image, it failed to encapsulate the osteophyte point locations (given as per the dataset). 
This is most likely because the reader was asked to place the marker along the center of gravity for each vertebra.

Therefore, a simple contour of the vertebra  did not suffice. To account for \final{the inclusion of the osteophyte location into the final resulting patches}, another strategy was implemented consisting of expanding the formed contour. 
The majority of the osteophytes occurred on the left side of the vertebrae hence we first expanded the contour in the -X direction. To account for the curvature of the spines, we also expanded the contour in the +Y direction. This allowed for the gaps between the vertebrae to be covered. Different size of expansion was utilized for cervical and lumbar scans to account for the variation in vertebra sizes. SegPatch finally resulted in 1680 present patches and 1517 absent patches, which was more favourably distributed than the tiling method described above. The details and visual depiction are outlined in Fig.~\ref{fig:pipeline}. \final{A rigid rule and clinical prior based heuristic was used to expand the contour to prevent further processing of the image prior to being fed to the classifier.}

\begin{figure}
    \centering
    \includegraphics[width=0.48\textwidth]{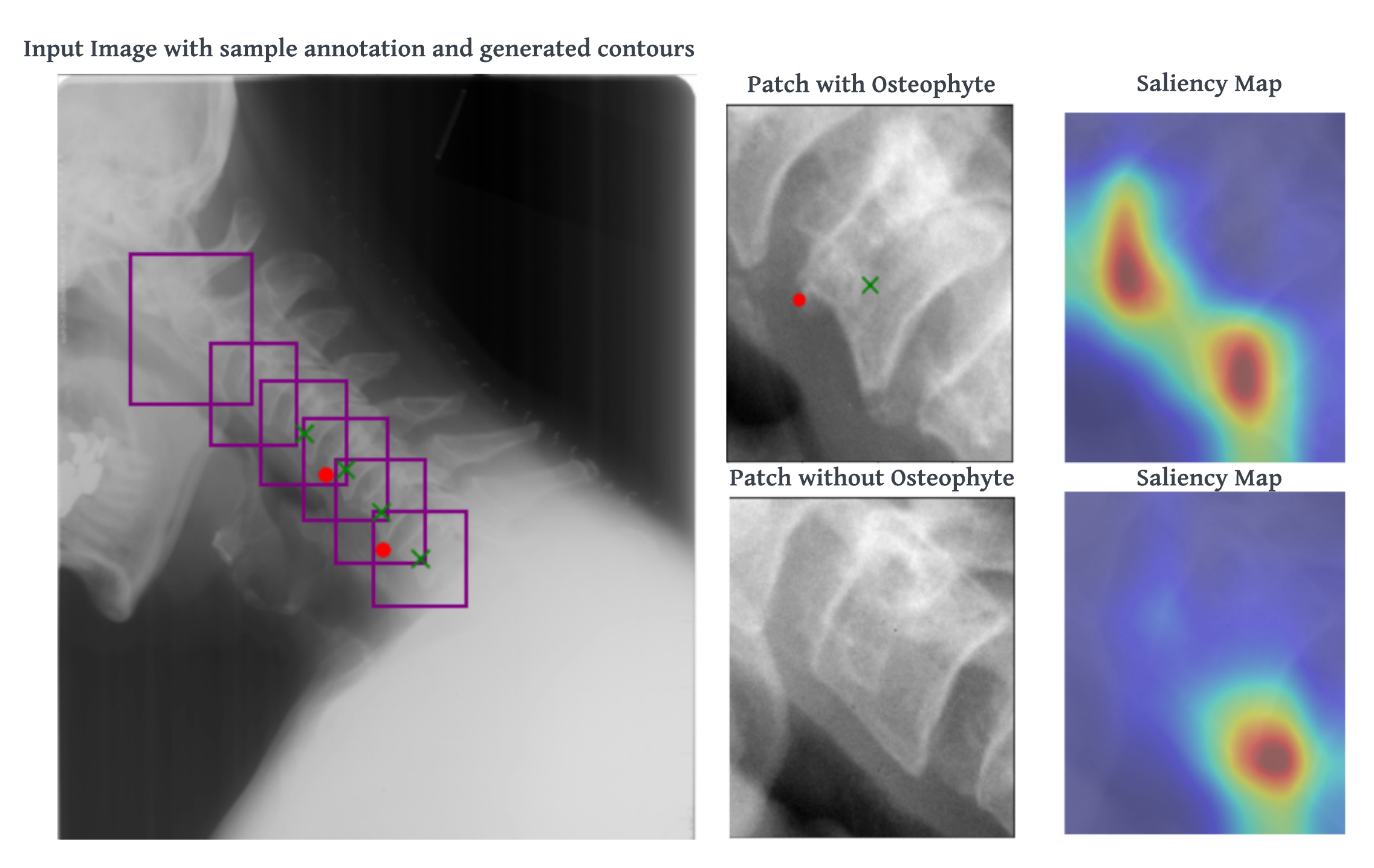}
    \caption{Saliency results for a positive and a negative patch. Red dots: Osteophytes. Purple boxes: Contours.}
    \label{fig:Explainability}
\end{figure}

\subsection{Classification of Osteophytes}
\label{sec:patch_classification}

Multiple state-of-the-art classifiers were tested including ResNet-18/50 and DenseNet-121/169, all pretrained on ImageNet.
Given the limited amount of annotated images, we employed data augmentation including random rotations (up to 30 degrees) and random histogram equalization (with a variance of 0.5). 
The input dimensions for both strategies are the same. The image patches were resized to (244 x 224) for SegPatch, and the tiling limit was set to (244 x 224).
For both methods, the cross-entropy loss, and the Stochastic Gradient Descent were used. 
Tiling was fine-tuned with a learning rate of 0.01 and a momentum of 0.7, while SegPatch achieved optimal results with a learning rate of 0.002 and a momentum of 0.09.
Additionally, a learning rate scheduler with a step size of 7 and a gamma value of 0.1 was employed. Peak performance was reached at 50 epochs for both approaches.




    

    



\section{Results}

As indicated in Tab.~\ref{Segpatch_Scores}, the tiling method produced an average test accuracy of 75.4\%, whereas SegPatch surpasses this performance, achieving an average accuracy of 84.5\%. 

Furthermore, Fig.~\ref{fig:Explainability} reveals an interesting observation: in the present patch, the saliency maps generated using SS-CAM are predominantly concentrated over the edges where the osteophyte is present. In the absence of an osteophyte, it primarily focuses on the edge but not on the location of the osteophyte. This observation vividly illustrates that the model is discerning the edges and making determinations regarding the presence or absence of growth by looking specifically at the edges of the vertebrae.

\begin{table}[]
\centering
\caption{Classification results for the proposed patching algorithms.}
\label{Segpatch_Scores}
\begin{tabular}{|lcccc|}
\hline
\multicolumn{1}{|c|}{\multirow{3}{*}{\textbf{Model}}} & \multicolumn{4}{c|}{\textbf{Accuracy Score}}                                                           \\ \cline{2-5} 
\multicolumn{1}{|c|}{}                                & \multicolumn{2}{c|}{\textbf{SegPatch}}                           & \multicolumn{2}{c|}{\textbf{Tiling}} \\ \cline{2-5} 
\multicolumn{1}{|c|}{}                                & \multicolumn{1}{c|}{Train}          & \multicolumn{1}{c|}{Test}  & \multicolumn{1}{c|}{Train}  & Test   \\ \hline
\multicolumn{1}{|l|}{ResNet-18}                       & \multicolumn{1}{c|}{84.80}           & \multicolumn{1}{c|}{80.34} & \multicolumn{1}{c|}{80.10}     & 73.34  \\ \hline
\multicolumn{1}{|l|}{ResNet-50}                       & \multicolumn{1}{c|}{85.34} & \multicolumn{1}{c|}{82.00}    & \multicolumn{1}{c|}{79.80}   & 75.40   \\ \hline
\multicolumn{1}{|l|}{DenseNet-121}                    & \multicolumn{1}{c|}{87.05}          & \multicolumn{1}{c|}{\textbf{84.64}} & \multicolumn{1}{c|}{77.73}  & 70.05  \\ \hline
\multicolumn{1}{|l|}{DenseNet-169}                    & \multicolumn{1}{c|}{87.20}           & \multicolumn{1}{c|}{82.75} & \multicolumn{1}{c|}{76.90}   & 69.70   \\ \hline
\multicolumn{5}{c}{\small{Scores are averaged over 3 runs. All scores at $\pm$ 0.75.}}                            
\end{tabular}
\end{table}

\section{Discussion}
This paper introduces an automated approach for the detection of osteophytes in spinal X-rays. 
The method employs a specialized automated patch generation technique called SegPatch to generate input data for a patch classifier, which is built upon a fine-tuned DenseNet-121 model. 
The method secured an overall 84.5\% accuracy score and compares favorably with the baseline tiling method.

\subsection{Solving the class imbalance in Tiling}
\final{Using rank-consistent ordinal regression framework (CORN) loss \cite{Shi_2023} or focal loss \cite{lin2017focal} to tackle class imbalance did not show any improvement over weighted cross-entropy. Neither did the use of over-sampling. Choosing a fixed number of random patches runs the risk of missing the already diminished number of positive patches. Owing to the significant variability in the images, an algorithmic strategy to eliminate a larger number of negative sample seemed infeasible. Experimentation also involved utilising larger models for the patch classification task but they exhibited over-fitting and were consequently excluded from consideration.}

\subsection{Problems with generating more precise patches}
The solution to class imbalance would be to generate more precise patches \final{focused on the edges of the vertebrae}. However, generating more precised patches faces quite a few challenges. The primary problem arises with the extreme curvature of the spine in certain scans. 

\begin{figure}
    \centering
    \includegraphics[width=0.32\textwidth]{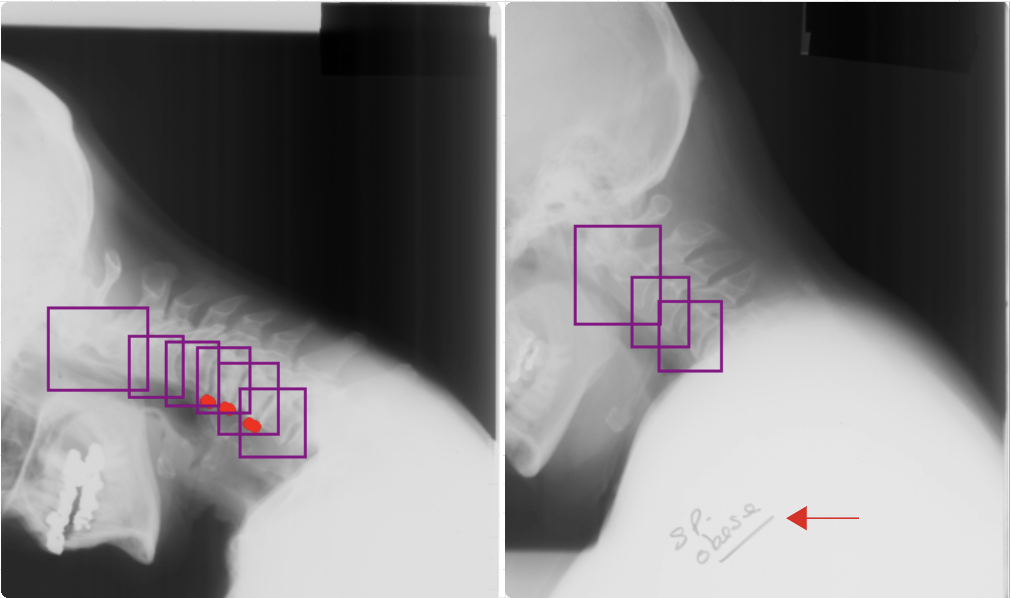}
    \caption{Example of difficult (but successful) to label scans. (Left): High curvature of spine. (Right): Presence of artifact  (text) denoted by red arrow with  with no osteophyte labels.}
    \label{fig:difficult}
\end{figure}

There are also cases with artifacts present on the scan, e.g. text on the scan (see red arrow in Fig. \ref{fig:difficult}). Given the random nature of occurrence of osteophytes, one has to account for each corner of each vertebrae in every scan. 
There is also a size difference in the Lumbar and Cervical scans. Hence, methods have to be size-invariant like that of SegPatch.

\begin{figure}
    \centering
    \includegraphics[width=0.25\textwidth]{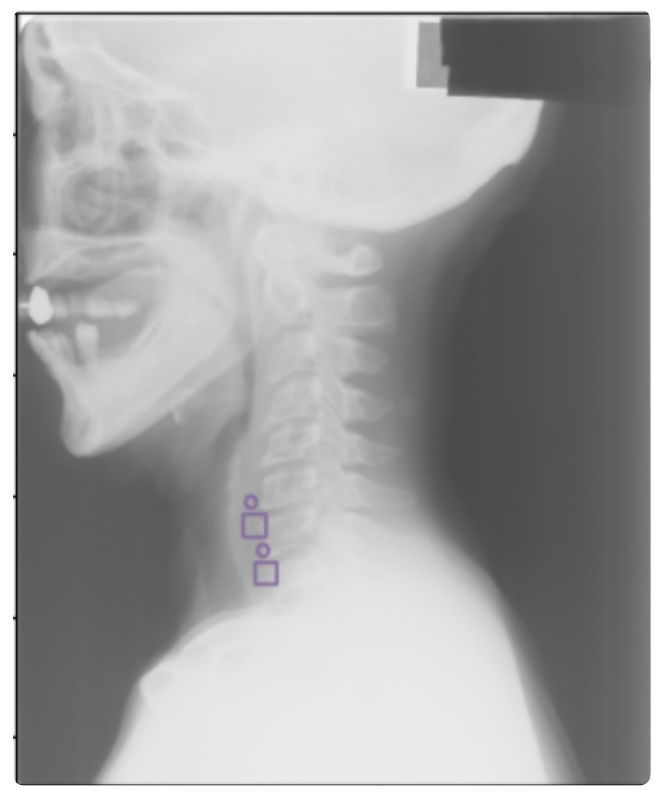}
    \caption{Visualisation of poor performance from off-the-shelf detector - FasterRCNN. Purple circles denote annotation location and squares denote predicted bounding box.}
    \label{fig:detection-results}
\end{figure} 

\subsubsection{Patch size variability in SegPatch} 
While tiling produces a uniform patch size, SegPatch generates non-uniformly sized patches which may lead to lower resolution patches for the larger patches generated. 
Uniform patch size was not feasible as that either resulted in multiple missing osteophytes or a large number (and area) of overlaps between the patches. 




\subsection{Poor Performance from off-the-shelf detectors}
\label{detection}
 Initially, to develop a solution for osteophyte detection, we investigated the state-of-the-art detection method, FasterRCNN \cite{girshick2014rich}, and achieved an average mAP and accuracy score of 0.1\%, 0.2\%, 20\% and 30\% as the size of the bounding increasing from 10, 18, 36 to 54 pixels, respectively.
 The results did not improve with different forms of augmentation (including tiling, which is originally devised for detection-based tasks) or utilizing the latest versions of YOLO (NAS, v8). The key positive insight derived from the detection task using FasterRCNN is seen in Fig. \ref{fig:detection-results}. The model clearly understands the typical locations where the osteophytes appear.
 Notably at the corners of the vertebrae. A trend similarly observed in VinDr-SpineXR study \cite{nguyen2021vindr}. Consequently, it is advisable for future methodologies to place greater emphasis on the identification aspect rather than localization in the context of osteophyte detection.

\section{Conclusion}


This paper presents one of the first studies for spinal osteophyte detection. It integrates the use of a specialized patching strategy called SegPatch in conjunction with a fine-tuned DenseNet-121 network. By employing patch classification, the identification process gains enhanced efficiency in detecting osteophytes as osteophytes tend to generally originate in the corner regions of vertebrae. Our method achieves 84.5\% accuracy score and a 86.62\% specificity score for the patch classifier which generates saliency maps that align well with both positive and negative classifications. A study done by Nevalainen et al., showed a specificity of 75\% in detecting osteophytes from MRI \cite{nevalainen2023ultrasound}. 

The findings, combined with the analysis of the results in this paper lay the groundwork for a more effective clinical approach to osteophyte detection in X-ray images and other smaller structures. 
\final{This is all while being robust to multiple orientations of the spine present in the scans, artifacts, minimal annotations and comparatively small training data.}

In addition, this work shows that osteophytes can be detected using a single modality instead of multi-modality approaches \cite{wang2016detection}.
This efficiency not only saves valuable time but also has the potential to enhance the accuracy and consistency of diagnoses, ultimately improving patient care and treatment outcomes.



\printendnotes

\subsection*{Acknowledgments}
S. S. Kundu was supported by the UK Medical Research Council
[MR/N013700/1] and the King’s College London MRC Doctoral Training Partnership
in Biomedical Sciences. Y. Mo is supported by the Oxford BDI-Novartis collaboration for AI in medicine. B. W. Papiez acknowledges Senior Fellowship at Population Health. The computational aspects of this research were supported by the Wellcome
Trust Core Award Grant Number 203141 /Z/16/Z and the NIHR Oxford BRC. The authors would also like to thank Mona Furukawa for her edits on the final manuscript.



\bibliographystyle{IEEEbib}
\bibliography{ref}

\end{document}